# Alleviation of Temperature Variation Induced Accuracy Degradation in Ferroelectric FinFET Based Neural Network

Sourav De[1,2*], Hoang-Hiep Le[2], Md. Aftab Baig[2], Yao-Jen Lee[3], Darsen D. Lu[2*] and Thomas Kämpfe[1]

1. Fraunhofer-Institut für Photonische Mikrosysteme IPMS - Center Nanoelectronic Technologies; sourav.de@ipms.fraunhofer.de
2. Institute of Microelectronics, National Cheng Kung University; darsenlu@mail.ncku.edu.tw
3. Institute of Pioneer Semiconductor Innovation, National Yang-Ming Chiao Tung University; yjlee1976@gmail.com
* Correspondence: sourav.de@ipms.fraunhofer.de

**Abstract:** This paper reports the impacts of temperature variation on the inference accuracy of pre-trained all-ferroelectric FinFET deep neural networks, along with plausible design techniques to abate these impacts. We adopted a pre-trained artificial neural network (N.N.) with 96.4% inference accuracy on the MNIST dataset as the baseline. As an aftermath of temperature change, a compact model captured the conductance drift of a programmed cell over a wide range of gate biases. We observed a significant inference accuracy degradation in the analog neural network at 233 K for an N.N. trained at 300 K. Finally, we deployed binary neural networks with "read voltage" optimization to ensure immunity of N.N. to accuracy degradation under temperature variation, maintaining an inference accuracy of 96.1%.

## 1. Introduction

The invention of the convolutional neural network [1] (CNN) changed the paradigm of computing and made machine learning or artificial intelligence a *'de-facto'* choice for solving many computationally challenging problems [2-4]. Recent advents in the research of hafnium oxide (*HfO₂*) based ferroelectric (Fe) materials have paved the way for using ferroelectric memories like ferroelectric field effect transistors (Fe-FETs) and ferroelectric random access memories (Fe-RAM) for the next generation memories, especially for computing in memory applications (CIM). Pronounced ferroelectricity in a single-layer thin film of $HfO_2$ [5-11], fast switching, high on-current ($I_{ON}$) to off-current ($I_{OFF}$) ratio $\frac{I_{ON}}{I_{OFF}}$ [12-16]. The critical technological factors that make Fe-FETs superior to other emerging non-volatile memory (eNVMs) technologies are excellent linearity, bi-directional operation, and good endurance [17-22]. Nevertheless, a significant temperature-induced threshold voltage ($V_{th}$) shift in Fe-FETs poses a challenge for practical applications [18,21,22]. Despite the random variation-tolerant nature of deep neural networks (DNN), the CIM architecture, where the analog current sum determines computational output, is susceptible to systematic variation in its basic device building block. This work's cornerstone is to mitigate the impact of temperature-change-induced variations in Fe-FET-based CIM operation without affecting device scaling, power consumption, and latency. The choice of Fe-FinFET as synaptic devices instead of planar FE-FET ensures compatibility with the perpetual device scaling trend. The paper discusses the fabrication and characterization of deeply scaled Fe-FinFETs, where we characterize and construct compact models with temperature variation effects. Subsequently, we investigate DNN applications focusing on offline training, where pre-trained weights are programmed to Fe-FinFET devices. The **(ii)** model is subsequently applied to a DNN, pre-trained with the CIMulator [23,24] software platform to evaluate inference accuracy degradation with temperature decrease. Based on the findings, we propose a novel application strategy to minimize the impacts of temperature variation.

## 2. Materials and Methods

We have fabricated Fe-FinFET using a self-aligned gate-first process on silicon-on-insulator (SOI) using the process described in [26, 15, 6, 4], with a 5 nm thin hafnium zirconium oxide (HZO) layer. Fig. 1(a) shows the schematic illustration, and Fig. 1(b) shows the transmission electron microscopic (TEM) cross-section of the fabricated Fe-FinFET with 15 nm top fin width ($T_{fin}$) and 5.8 nm ferroelectric layer ($T_{FE}$).

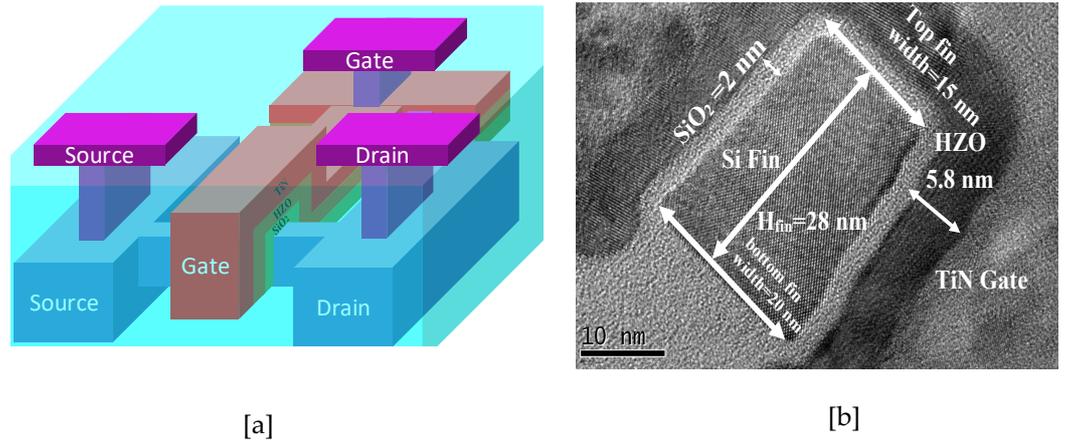

[a] [b]

Fig. 1. (a) Schematic illustration of ferroelectric finFET devices. (b) TEM cross-section of the fabricated device. The average fin width is 18 nm, and the fin height is 28nm. The thickness of the F.E. layer (HZO) is 5.8 nm.

The electrical characterization was done using 100 ns pulse of amplitude 4.5V and -4.5V respectively at the gate terminal for binary program-erase operations. The drain terminal was kept at 0V during the program operation, and the read operation was performed by applying a non-disturbing dc sweep from -0.5 V to 1.5 V at gate voltage while keeping the drain terminal at a constant 100mV. The multilevel characteristics are also evaluated by partially switching the ferroelectric layer's dipoles by fast and low voltage pulses. 100ns-wide pulses with constant amplitude (2V) are applied for programming (LTP); pulses of opposite polarity (-2V) are applied for erasing (LTD). The resulting remnant polarization alters the inversion charge concentration in the channel, modulation in channel conductance, and the threshold voltage.

The program-erase operation was followed by analyzing temperature variation on synaptic characteristics of Fe-finFET devices. Temperature change in real scenarios causes mobility and carrier concentration to fluctuate, leading to change in $G_{ch}$ and $V_{th}$ for Fe-FinFET synaptic devices, thus altering the N.N. weights they represent. The temperature dependence of the FE-FinFET is characterized by first programming the device to a fixed (low or high resistance) state at room temperature (300 K), measuring $I_d$-$V_{gs}$ with a non-destructive $V_{gs}$ range, reducing the temperature, and measuring again.

## 3. Results

The binary program-erase operation is shown in Fig. 2(a). Although the memory window in-terms-of threshold voltage shift is only 500mV, the devices show a superior $\frac{I_{ON}}{I_{OFF}}$ of $10^6$ with an on-state current of 100nA. Figure 2(b) depicts the channel conductance modulation and multilevel states measured from a single Fe-FinFET cell with 20nm $T_{fin}$ and 50nm gate length ($L_g$). The device shows excellent linearity and symmetry with a nonlinearity factor of -0.84 for long-term potentiation (LTP) and long-term depression (LTD).

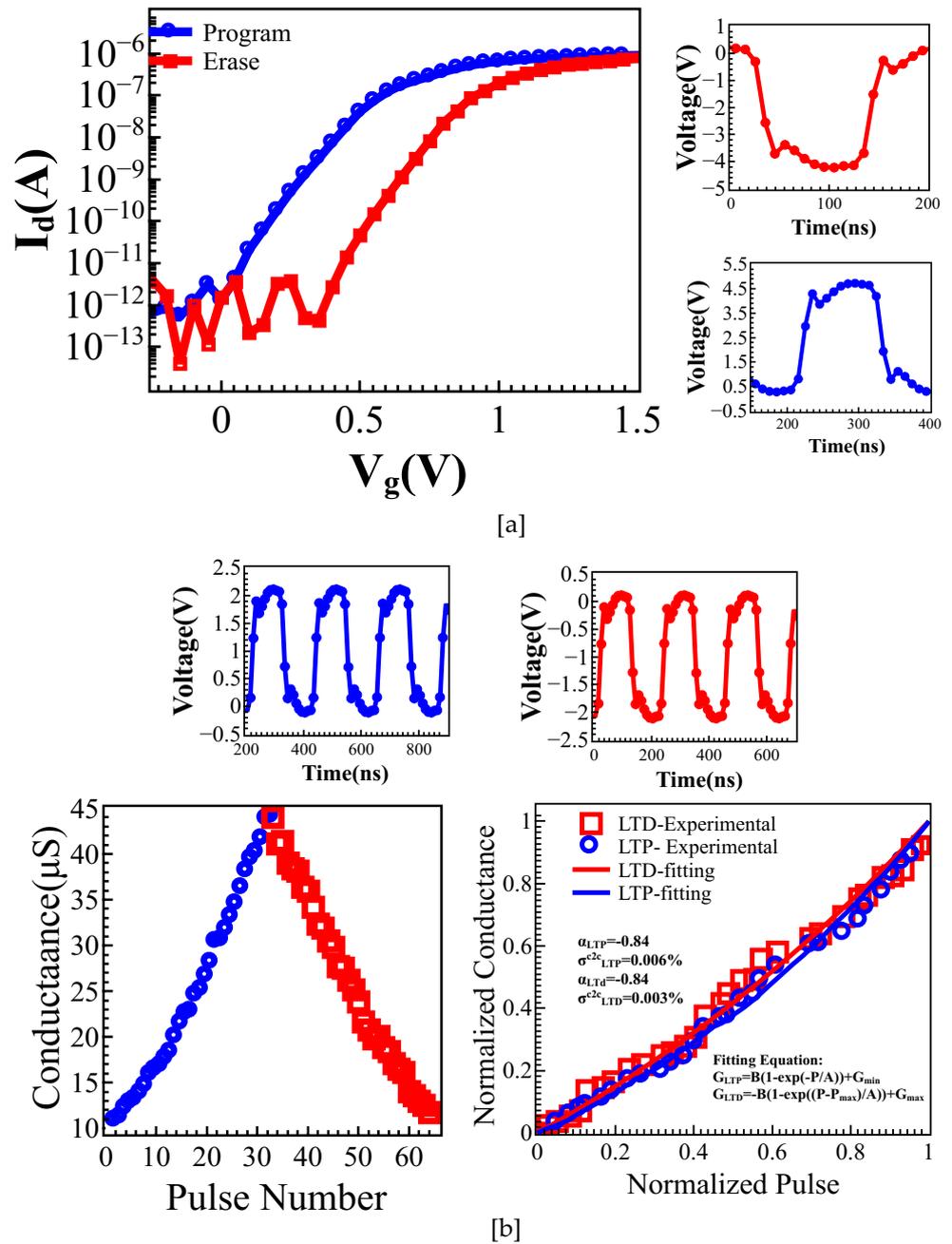

Fig. 2. (a) Binary program-erase characteristics. The low resistance state has been obtained by programming the device with a 4.5V pulse of 100ns width. The high resistance state has been obtained with a -4.5V pulse of 100ns width. (b) The gradual change of channel conductance during the analog program and erase operations were obtained by applying a ±2V pulse of 100ns at the gate terminal. The drain terminals were kept at 0V during the program and erase operations. The read operations are similar to the technique described in [4,23]. A constant 100mV was applied at the drain terminal during the read operation. The channel conductance was extracted at a constant 1V gate voltage, which shows excellent linear and symmetric characteristics. The nonlinearity factor has been calculated similarly described in [4,18].

The impacts of temperature on mobility and threshold voltage are modeled with the following expressions [23]:

$$\mu(T) = \mu(300K)\left(\frac{T}{300}\right)^{UTE} \quad (1)$$

$$V_{th}(T) = V_{th}(300) + KT1\left(\frac{T}{300} - 1\right) \quad (2)$$

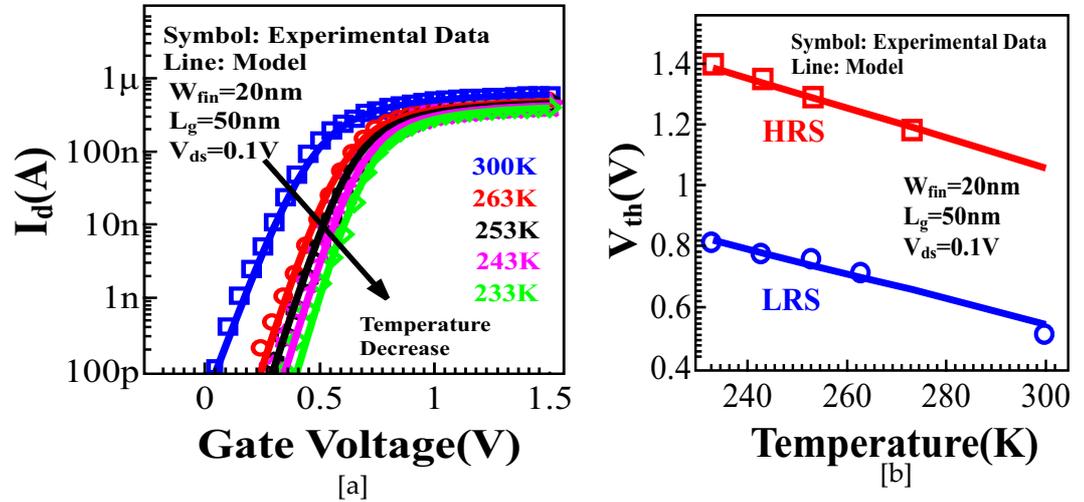

Fig. 3. (a) Characterization and modeling of the impact of temperature change for the low resistance state. (b) Characterization and modeling the temperature change induced threshold voltage shift in FE-FinFETs.

The evolution of Vth with the change of temperature for the low-resistance state (LRS, programmed using a +4.5V pulse) is accurately captured (Fig. 3(b)). The same expressions are used for the high-resistance state (HRS), with good agreement with measured data. Fig. 4 shows the measured and modeled change in Gch as a function of temperature, which highlights the dependence of the ON/OFF ratio on $V_{g,read}$. It is evident that to ensure a high ON/OFF ratio, the device should be in the subthreshold region at all temperatures in HRS, which is key to the proper operation of binarized neural networks (BNN).

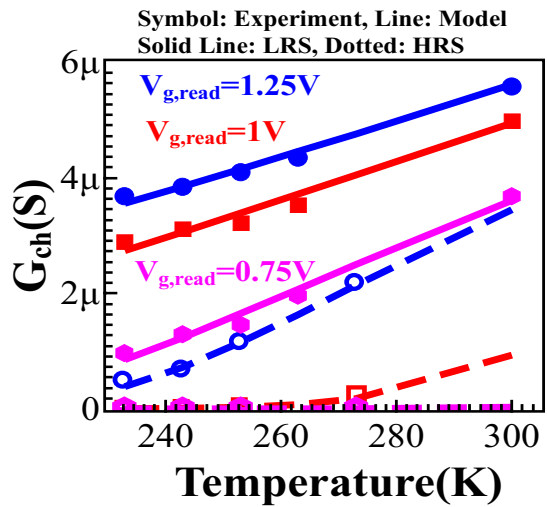

Fig. 4. Characterization and modeling the conductance drift as a result of temperature change for different read voltage.

We further applied the calibrated model to consider a realistic scenario (Fig. 5 (a)), where BNN was trained at room temperature and driven to a lower temperature to perform inference without re-calibration. When a temperature change occurs, we observe a significant drop in N.N. accuracy (10%) for $V_{g,read}$=1.25V, because of the significant

ON/OFF ratio change when the operation regions of HRS synapses go from strong inversion at 300K to weak inversion at 233K. Although one of our previous works [22] shows that re-calibration at 233K may restore accuracy, the overhead associated with it may not be suitable for real applications. On the other hand, $V_{g,read}$ of 0.75V, ensures the requisite ON/OFF ratio (HRS stays in the sub-threshold region). The inference accuracy remains unfazed amidst temperature change only when $V_{g,read}$ is 0.75V. For all other cases, a significant accuracy drop is noted with at least one state switching from strong inversion to sub-threshold (Fig.5 (b)). The aforementioned $V_{g,read}$ optimization ensures preserving the digital nature (high ON/OFF ratio) of N.N. weights. Accurate modeling of $G_{ch}$, ranging from sub-threshold to strong inversion, is crucial for optimizing bias conditions to ensure proper N.N. operation.

The results of this study lead to a possible "dynamic read voltage" scheme, where $V_{g,read}$ is chosen according to the present junction temperature. We expect such a scheme to alleviate the impact of temperature variation and expand the overall design margin.

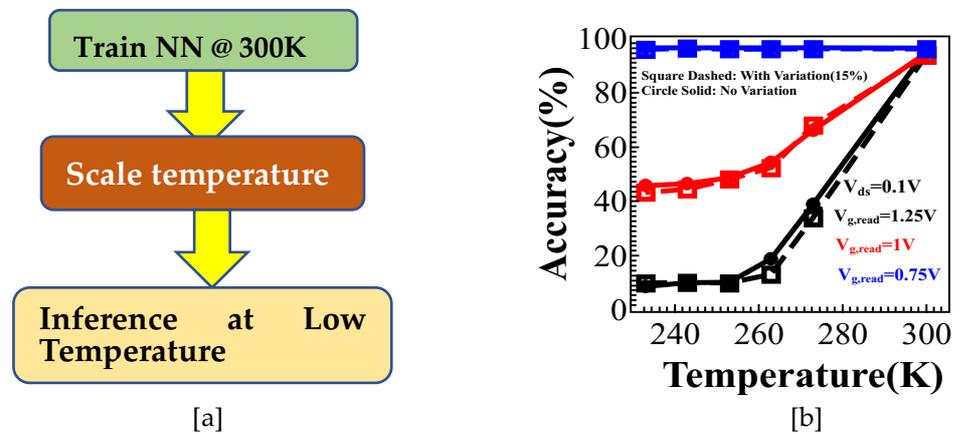

Fig.5. (a). Flow-chart of neuromorphic simulation for obtaining inference at a lower temperature after training at room temperature. (b). The optimal choice of read voltage is required to avoid accuracy degradation due to temperature-induced conductance drift.

### 4. Discussion

We have fabricated, characterized, and evaluated the performance of deeply scaled FE-FinFETs for neuromorphic computing in the presence of temperature variation. The digital nature of the binarized neural network, with the "0" state programmed deep in sub-threshold and "1" state in strong inversion, is key to robust DNN inference. On the other hand, the apparent channel conductance shift of FE-FET as an aftermath of temperature change may render the neural network inaccurate due to its analog nature. A proper choice of "read voltage" is paramount to ensure the "0" state stays deep in the sub-threshold region for maintaining the ON/OFF ratio for the LRS-HRS state. This study also suggested a dynamic read voltage scheme to maintain inference accuracy in the presence of temperature variation.

**Supplementary Materials:** The following supporting information can be downloaded at:

**Author Contributions:** Conceptualization, Sourav De, Thomas Kämpfe and Darsen Lu; methodology, Sourav De, Yao-Jen Lee and Darsen Lu; software, Hoang-Hiep Le and Md.Aftab Baig; validation, Sourav De, and Darsen Lu; formal analysis, Sourav De; investigation, Sourav De.; resources, Yao-Jen Lee; data curation,Darsen Lu and Thomas Kämpfe ; writing—original draft preparation, Sourav De; writing—review and editing, Darsen Lu and Thomas Kämpfe; visualization, Sourav De; supervision, Darsen Lu, Yao-Jen Lee and Thomas Kämpfe; project administration, Darsen Lu, Yao-Jen Lee and Thomas Kämpfe; funding acquisition, Darsen Lu, Yao-Jen Lee and Thomas Kämpfe. All authors have read and agreed to the published version of the manuscript."

**Funding:** This work was funded by Ministry of Science and Technology grants MOST-108-2634-F-006-008 and MOST-109-2628-E-492 -001 -MY3 and also this research was funded by the ECSEL Joint Undertaking project ANDANTE in collaboration with the European Union's Horizon 2020 Framework Program for Research and Innovation (H2020/2014-2020) and National Authorities, under Grant No. 876925.

**Data Availability Statement:** data is not available.

**Acknowledgments:** We are grateful to TSRI for providing us with advance CMOS fabrication platform, and NCHC for GPU computing facilities.

**Conflicts of Interest:** "The authors declare no conflict of interest. The funders had no role in the design of the study; in the collection, analyses, or interpretation of data; in the writing of the manuscript; or in the decision to publish the results".